\documentclass{article}
\usepackage{spconf,amsmath,graphicx}

\usepackage{cite}
\usepackage{graphicx}
\usepackage{float}
\usepackage{amsmath}
\usepackage{multirow}
\usepackage{longtable}
\usepackage{array}
\usepackage{tcolorbox}
\usepackage{booktabs}
\usepackage{tikz}
\usepackage{framed}
\usepackage{wrapfig}
\usepackage{pifont}
\usepackage{url}
\usepackage[hidelinks]{hyperref}
\usepackage{amssymb}    
\usepackage{amsfonts}   

\title{Persuasion Should be Double-Blind: A Multi-Domain Dialogue Dataset With Faithfulness Based on Causal Theory of Mind}
\name{\bf{Dingyi Zhang}$^{1}$, 
      \bf{Linhai Zhang}$^{2}$, 
      \bf{Fanglei Qu}$^{1}$, 
      \bf{Ziqing Zhuang}$^{1}$,
      \bf{Deyu Zhou}$^{1 *}$\thanks{* Corresponding author (d.zhou@seu.edu.cn).}
      \thanks{The CToMPersu dataset, together with the associated prompts and code, is available at \href{https://github.com/DingyiZhang/MAP-A-Meta-Cognitive-Autonomous-Intelligent-Agents-Framework-for-Complex-Persuasion}{GitHub Repo}}}
\address{
$^{1}$School of Computer Science and Engineering, Key Laboratory of Computer Network \\ and Information Integration, Ministry of Education, Southeast University, China \\
$^{2}$Department of Informatics, King’s College London\\
}

\begin{document}
\ninept
\maketitle
\begin{abstract}
Persuasive dialogue is central to human communication, yet existing datasets often rely on a single language model generating both roles, producing unrealistic interactions that violate the double-blind nature of persuasion. To overcome this, we propose ToMMA, a multi-agent framework guided by causal Theory of Mind that enforces role separation and prevents information leakage. Using ToMMA, we build CToMPersu, a large-scale multi-turn, multi-domain dataset capturing realistic persuasion dynamics. Automatic evaluations show that CToMPersu produces more coherent and persuasive dialogues than prior datasets. Furthermore, when used as a knowledge base, CToMPersu significantly enhances the persuasive performance of large language models, as confirmed by both automatic and human evaluations.
\end{abstract}

\section{Introduction}
\label{sec:intro}

Persuasive dialogue underpins applications in education, healthcare counseling, and business marketing \cite{rogiers2024persuasion}. High-quality datasets are therefore essential for developing and evaluating such systems. Human-authored corpora collected from real interactions or crowd role-play provide naturalness but are typically small and domain-specific—e.g., charity donation \cite{wang-etal-2019-persuasion}, recommendation \cite{10.5555/3327546.3327641}, and medical consultation \cite{zeng-etal-2020-meddialog}—which limits generalizable modeling. This authenticity–breadth trade-off remains a central challenge for dataset-driven research in persuasive dialogue.

Two dataset paradigms dominate persuasive dialogue research: human-authored corpora, which offer realism but remain small and domain-narrow; and LLM-generated datasets, where a single model produces both persuader and persuadee via prompting \cite{jin-etal-2024-persuading}. The latter greatly improves scalability and domain coverage, yet the absence of role separation induces information leakage and violates the double-blind nature of persuasion; moreover, without explicit Theory-of-Mind reasoning, persuaders often fail to address the persuadee’s underlying beliefs and desires, reducing logical consistency and realism. Two failure modes exemplify the problem. First, double-blind leakage: for example, a persuadee who prefers popular attractions abruptly cites “crowds,” mirroring the persuader’s pre-assigned “avoid tourist areas” strategy and thereby weakening their own stance. Second, Theory-of-Mind misalignment: for example, when the persuadee’s belief is “kids enjoy rides” and the desire is “have a happy vacation,” the persuader instead promotes the Great Wall for “family bonding,” addressing neither belief nor desire. Thus, despite breadth, current LLM-based datasets fall short of capturing human-like persuasive dynamics.

Our work makes three main contributions: 
(i) we propose a causal Theory-of-Mind evaluation that checks belief–desire satisfaction and enforces reasoning consistency; 
(ii) we introduce \textbf{ToMMA}, a role-separated multi-agent framework that enables double-blind, ToM-consistent persuasive dialogue generation; 
and (iii) we build \textbf{CToMPersu}, a large-scale multi-domain dataset validated by both automatic and human evaluations.

\section{Related Work}
\noindent\textbf{Persuasion}\quad
LLM-based persuasion has been explored in health, politics, and product recommendation \cite{altay2023information, potter-etal-2024-hidden, chen2023would}, as well as personalization and personality effects \cite{lou2025personalitymodelingpersuasionmisinformation, ju2025adaptivepsychologicalpersuasionlarge}. Beyond text-based studies, persuasion has also been investigated from a speech and acoustic perspective. Recent ICASSP work has explored persuasion-related phenomena, including the detection of check-worthy claims in political speech and the influence of vocal attributes such as loudness and smiling on timbre and perceived charisma \cite{ivanov2024checkworthy, shi2024charismatic}. Other studies examined strategy detection or scoring frameworks \cite{jin-etal-2023-joint, saenger-etal-2024-autopersuade}. Regarding datasets, PersuasionForGood and MedDialog provide small-scale human dialogues \cite{wang-etal-2019-persuasion, zeng-etal-2020-meddialog}, while large-scale multi-domain corpora such as DailyPersuasion expand coverage but struggle with consistency \cite{jin-etal-2024-persuading}.

\vspace{0.8em}\noindent\textbf{Theory of Mind}\quad
Theory of Mind (ToM) refers to attributing beliefs and desires to others \cite{premack1978does}, modeled through frameworks such as BDI and causal ToM \cite{georgeff1999belief, wu2024change}, and assessed by false belief tasks \cite{baron1985does}. Inspired by psychology, NLP studies test ToM in LLMs via benchmarks such as ToMi, FANToM, and ToMBench \cite{le-etal-2019-revisiting, kim-etal-2023-fantom, chen-etal-2024-tombench}. Multi-agent studies further exploit ToM in tasks and games \cite{yim2024evaluating, cross2024hypothetical}, and planning-oriented evaluations test whether LLMs reason about future actions \cite{moore2025largelanguagemodelsplanning}. These works indicate strong potential for integrating ToM into multi-agent LLM dialogue systems.

\begin{figure*}[t]
\centering
\includegraphics[width=0.95\textwidth]{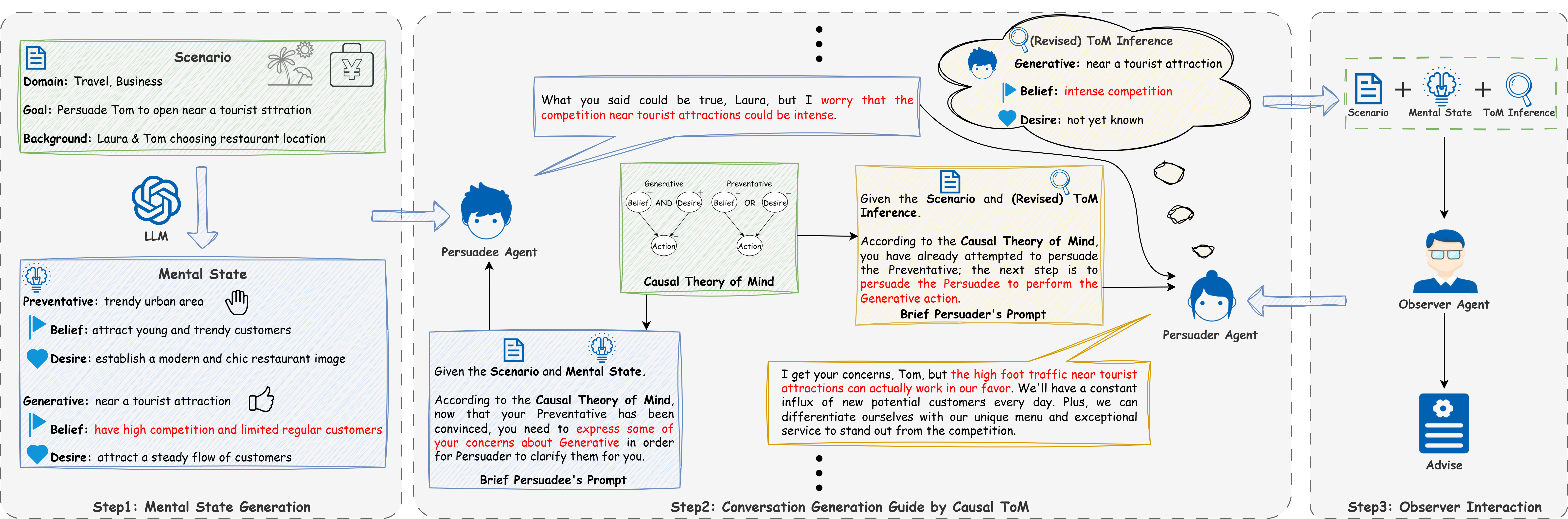} 
\caption{Overview of the \textbf{ToMMA} framework for collecting the \textbf{CToMPersu} dataset. This figure illustrates the three-step process: (1) Mental State Generation, (2) Dialogue Generation Guided by Causal Theory of Mind, and (3) Observer Interaction for quality control.} 
\label{fig:solution_overview}
\vspace{-0.2cm}
\end{figure*}

\section{Method}
\label{sec:Method}

\subsection{Causal Theory of Mind Evaluation}

\vspace{1em}\noindent\textbf{Causal Theory of Mind}
Causal Theory of Mind (CToM) refers to using Theory of Mind to shape others’ behaviors \cite{wu2024change}. It distinguishes two cases: the \emph{Preventative} case, where preventing an action requires altering either belief or desire (e.g., telling someone the post office is closed, or removing their need to send a letter); and the \emph{Generative} case, where promoting an action requires satisfying both belief and desire (e.g., they must believe the post office is open and also need to send a letter). This causal structure grounds our evaluation framework.

\vspace{1em}\noindent\textbf{Evaluation Protocol.}
\label{sec:CToM_Eval}
To quantitatively operationalize the Causal Theory of Mind (CToM) as an evaluation metric, we propose a three-stage prompting process using GPT-4o:

\begin{enumerate}
    \item \textbf{Mental State Generation:}  
    Given a dialogue $\mathcal{D}$, GPT-4o first infers the persuadee's underlying psychological states, including generative and preventative components:
    \begin{itemize}
        \item Generative Belief and Desire: $B_g$, $D_g$
        \item Preventative Belief and Desire (if present): $B_p$, $D_p$
    \end{itemize}

    \item \textbf{Condition Satisfaction Judgment:}  
    GPT-4o then evaluates whether the persuader addresses each inferred psychological state within dialogue $D$, assigning binary values:
    \[
    B_g', D_g', B_p', D_p' \in \{0,1\}
    \]
    where $1$ indicates that the persuader successfully addressed the state, and $0$ otherwise.  

    Intermediate generative score $G$ and preventative score $P$ are calculated as:
    \begin{equation}
    G = \begin{cases}
        1 & \text{if } B_g' = 1 \text{ and } D_g' = 1\\[0.2em]
        0 & \text{otherwise}
    \end{cases}
    \end{equation}
    \begin{equation}
    P = \begin{cases}
        1 & \text{if } B_p' = 1 \text{ or } D_p' = 1\\[0.2em]
        0 & \text{otherwise}
    \end{cases}
    \end{equation}

    \item \textbf{Final CToM Evaluation Score:}  
    We define the final CToM evaluation score for dialogue $D$ as:
    \begin{equation}
    \text{CToM}_{\text{score}}(D) = G \times P
    \end{equation}

    A dialogue receives a CToM evaluation score of $1$ only if it simultaneously promotes the target action and addresses at least one preventative component.
\end{enumerate}

Finally, the overall CToM score for an entire dataset consisting of $N$ dialogues is calculated as the ratio of dialogues receiving a CToM evaluation score of $1$:
\begin{equation}
\text{CToM}_{\text{eval}} = \frac{1}{N}\sum_{i=1}^{N}\text{CToM}_{\text{score}}(D_i)
\end{equation}

\subsection{ToMMA}
We propose \textbf{ToMMA}, a multi-agent framework that generates persuasive dialogues under \emph{double-blind} constraints and \emph{causal Theory of Mind} guidance. As shown in Figure~\ref{fig:solution_overview}, the pipeline comprises three stages: (i) filtering scenarios and generating the persuadee’s mental state; (ii) role-separated dialogue generation where the persuader \emph{cannot} access the persuadee’s internal state and must infer it over turns; and (iii) observer-based quality control that detects mismatches and provides corrective feedback.

Given a set of scenarios $\mathcal{S} = \{s_1, s_2, \dots, s_M\}$ from DailyPersuasion, we define the multi-agent process as:
\begin{equation}
\mathcal{D} = \{ (s_i, M_i, G_i, O_i) \mid s_i \in \mathcal{S} \}
\end{equation}
where $M_i$ denotes the generated mental state for $s_i$, $G_i$ the generated dialogue, and $O_i$ the observer feedback. The process ensures $M_i$ and $G_i$ satisfy double-blind constraints.

\vspace{1em}
\noindent\textbf{Mental State Generation.}
Given scenario $s_i$ with background $b_i$ and roles $r_i$, GPT-4o generates:
\begin{equation}
M_i = f_{\text{MS}}(b_i, r_i) = (B_g, D_g, B_p, D_p)
\end{equation}
where $f_{\text{MS}}$ denotes using an LLM with carefully designed prompts to generate mental states. $(B_g,D_g)$ are \emph{Generative} components and $(B_p,D_p)$ are \emph{Preventative} components (if absent, set $B_p=D_p=\text{None}$). The persuader has no access to $M_i$; it is used only to condition the persuadee’s behavior and to support evaluation/quality control.

\vspace{1em}
\noindent\textbf{Conversation Generation.}
Given mental state $M_i$, the conversation generation module produces:
\begin{equation}
G_i = f_{\text{CG}}(s_i, M_i, T_i)
\end{equation}
where $f_{\text{CG}}$ denotes using an LLM with prompting templates to generate dialogues, and $T_i$ is the allocated number of turns. All utterances follow a compact prompting template $\phi$:
\begin{equation}
G_i = \{u_1, u_2, \dots, u_{2 * T_i}\}, \quad u_j \in \phi
\end{equation}

\vspace{1em}\noindent\textbf{Persuadee Agent}\quad
observes the scenario and its own $M_i$ and responds according to the dialogue history.

\vspace{1em}\noindent\textbf{Persuader Agent}\quad
The persuader only has access to the scenario and must infer the persuadee’s mental state through interaction. By subtly probing in the early turns and leveraging conversation history, the persuader gradually models the persuadee’s beliefs and desires using Theory of Mind.

Once the persuader has developed an understanding of the persuadee’s mental state, they craft customized persuasive strategies. According to Causal Theory of Mind, when addressing Preventative Behavior, the persuader focuses on influencing the belief or desire that is more responsive to persuasion, depending on which aspect is easier to change. For Generative Behavior, the persuader must address both the belief and the desire in order to align with the persuadee's motivations and influence their decision. 

\vspace{1em}
\noindent\textbf{Observer Interaction.}  
The observer directly evaluates whether the persuader’s inferred mental state $\hat{M}_i$ matches the ground-truth $M_i$:
\begin{equation}
\delta_i = \mathbb{I}[\hat{M}_i = M_i]
\end{equation}
If $\delta_i = 0$, the observer generates feedback $O_i$ based on the discrepancy:
\begin{equation}
O_i = f_{\text{Obs}}(M_i, \hat{M}_i)
\end{equation}

\noindent The persuader updates the inferred mental state using this feedback:
\begin{equation}
\hat{M}_i' = \text{Update}(\hat{M}_i, O_i)
\end{equation}
Finally, the persuader generates a new dialogue based on the updated mental state:
\begin{equation}
G_i' = f_{\text{CG}}(s_i, \hat{M}_i', T_i)
\end{equation}
This process ensures that the final dialogue $G_i'$ is consistent with the corrected mental state $\hat{M}_i'$.

\subsection{CToMPersu}
Leveraging the ToMMA framework, we construct \textbf{CToMPersu}, a large-scale, multi-domain persuasive-dialogue dataset that reflects authentic human interactions under double-blind conditions. It contains 6{,}275 dialogues across 35 domains (e.g., travel, health, technology, business). Each scenario explicitly specifies the persuadee’s Generative and Preventative mental states—belief and desire components—and dialogues are produced with strict role separation so that the persuader and persuadee are simulated independently without access to each other’s internal information. The generation process enforces causal Theory-of-Mind consistency: the persuader infers and addresses the persuadee’s beliefs and desires when constructing arguments. Together, these design choices yield dialogues that align closely with human persuasion logic while offering scale and domain diversity.

\begin{table}[t]
\centering
\setlength{\tabcolsep}{6pt}
\renewcommand{\arraystretch}{1.2}
\caption{Comparison of persuasive dialogue datasets on key properties.
MT = Multi-turn Dialogue, HD = Human Dialogue, LS = Large scale, 
MD = Multi domain, IS = Information Separation, 
MAI = Multi-agent Interaction. \ding{51} indicates presence, \ding{55} indicates absence.}
\vspace{1em}
\begin{tabular}{lcccccc}
\hline
\textbf{Dataset} & \textbf{MT} & \textbf{HD} & \textbf{LS} & \textbf{MD} & \textbf{IS} & \textbf{MAI} \\
\hline
PersuForGood & \ding{51} & \ding{51} & \ding{55} & \ding{55} & \ding{51} & \ding{51} \\
DailyPersu   & \ding{51} & \ding{55} & \ding{51} & \ding{51} & \ding{55} & \ding{55} \\
CToMPersu    & \ding{51} & \ding{55} & \ding{51} & \ding{51} & \ding{51} & \ding{51} \\
\hline
\end{tabular}
\label{table:dataset_comparison}
\end{table}

\section{Experiments}
In the experimental section, we demonstrate how our dataset compares with other human and LLM-generated datasets using conventional evaluation methods, as well as evaluating its consistency on three respects: Causal ToM Evaluation, Direct Prompting Effectiveness, and Double-Blind Consistency.

\subsection{Dataset Evaluation}
To evaluate CToMPersu, we compare it with PersuasionForGood, a small-scale human dialogue dataset focused on donation scenarios, and DailyPersuasion, a large-scale, multi-domain, multi-turn dataset generated by GPT-4. As shown in Table~\ref{table:dataset_comparison}, CToMPersu not only scales effectively but also preserves the double-blind and cognitively faithful dynamics characteristic of real persuasive communication.

\vspace{1em}\noindent\textbf{Metrics}\quad
The evaluation is conducted using three key metrics: (1) \textbf{Direct Prompting}, where the LLM is prompted to simulate the persuadee and judge whether the dialogue is convincing; (2) \textbf{Causal Theory of Mind Evaluation}, which assesses whether the persuader's utterances properly address the persuadee’s mental state based on causal ToM reasoning (see Section~\ref{sec:CToM_Eval}); and (3) \textbf{Double-Blind Inconsistency}, which checks whether the persuadee displays access to the persuader’s internal strategy or knowledge, thereby violating the double-blind constraint. All metrics are computed as pass rates over sampled dialogue instances using GPT-3.5 as the evaluator.

The results presented in Table~\ref{table:evaluation_results} demonstrate that CToMPersu achieves strong performance across all three metrics. While DailyPersuasion achieves slightly higher scores in Direct Prompting, its Causal ToM and Double-Blind consistency are substantially lower. In the Causal ToM Eval results, CToMPersu exhibits a modest drop of 8.8 points compared to its Direct Prompting score, which is closely aligned with the 9.0-point drop observed in the human-authored dataset PersuasionForGood. In contrast, DailyPersuasion shows a much larger gap of 35.9 points. This similarity in drop-off patterns suggests that CToMPersu better preserves the underlying persuasive logic found in real human conversations.

\begin{table}[t]
\centering
\small
\renewcommand{\arraystretch}{1.1}
\setlength{\tabcolsep}{4pt}
\caption{Comparison of Direct Prompting (DP), Causal ToM Evaluation (CT), and Double-Blind Inconsistency (DB) across three persuasive dialogue datasets.}
\vspace{1em}
\resizebox{\linewidth}{!}{
\begin{tabular}{lccc}
\hline
\textbf{Metric} & \textbf{PersuForGood} & \textbf{DailyPersu} & \textbf{CToMPersu} \\ \hline
DP & 88.87 & 90.75 & \textbf{90.82} \\
CT & 79.87 & 54.80 & \textbf{82.02} \\
DB & 97.77 & 78.57 & \textbf{95.56} \\ \hline
\end{tabular}
}
\label{table:evaluation_results}
\end{table}

\subsection{Experimental Results}

\vspace{1em}\noindent\textbf{Setup:}
We constructed a test set consisting of 525 dialogues, while the remaining dialogues from the dataset were used as an external knowledge base to support in-context learning (ICL). We evaluated the persuasive capabilities of several large language models, including GPT-3.5, GPT-4o, DeepSeek-V3, Gemini-2.0-Flash, and Gemini-2.5-pro-preview. For each model, we compared the effectiveness of using different datasets (CToMPersu vs. DailyPersuasion) as ICL sources, examining how the quality of data influences

\begin{figure*}[ht]
    \centering
    \begin{minipage}[t]{0.48\textwidth}
        \centering
        \includegraphics[width=\textwidth]{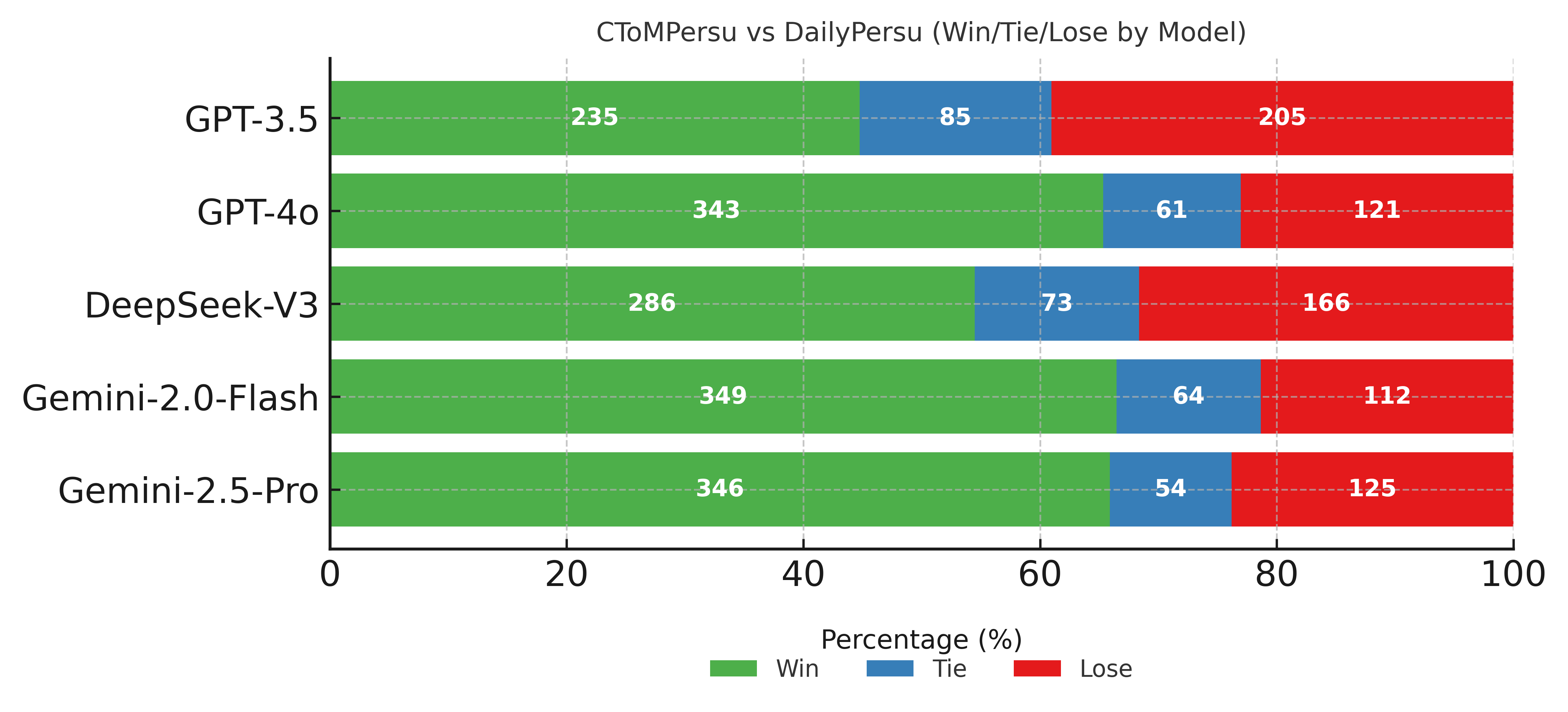}
        \vspace{0.3em}
        \small (a) LLM Evaluation Results: CToMPersu vs DailyPersuasion
        \label{fig:output_compare}
    \end{minipage}
    \hfill
    \begin{minipage}[t]{0.48\textwidth}
        \centering
        \includegraphics[width=\textwidth]{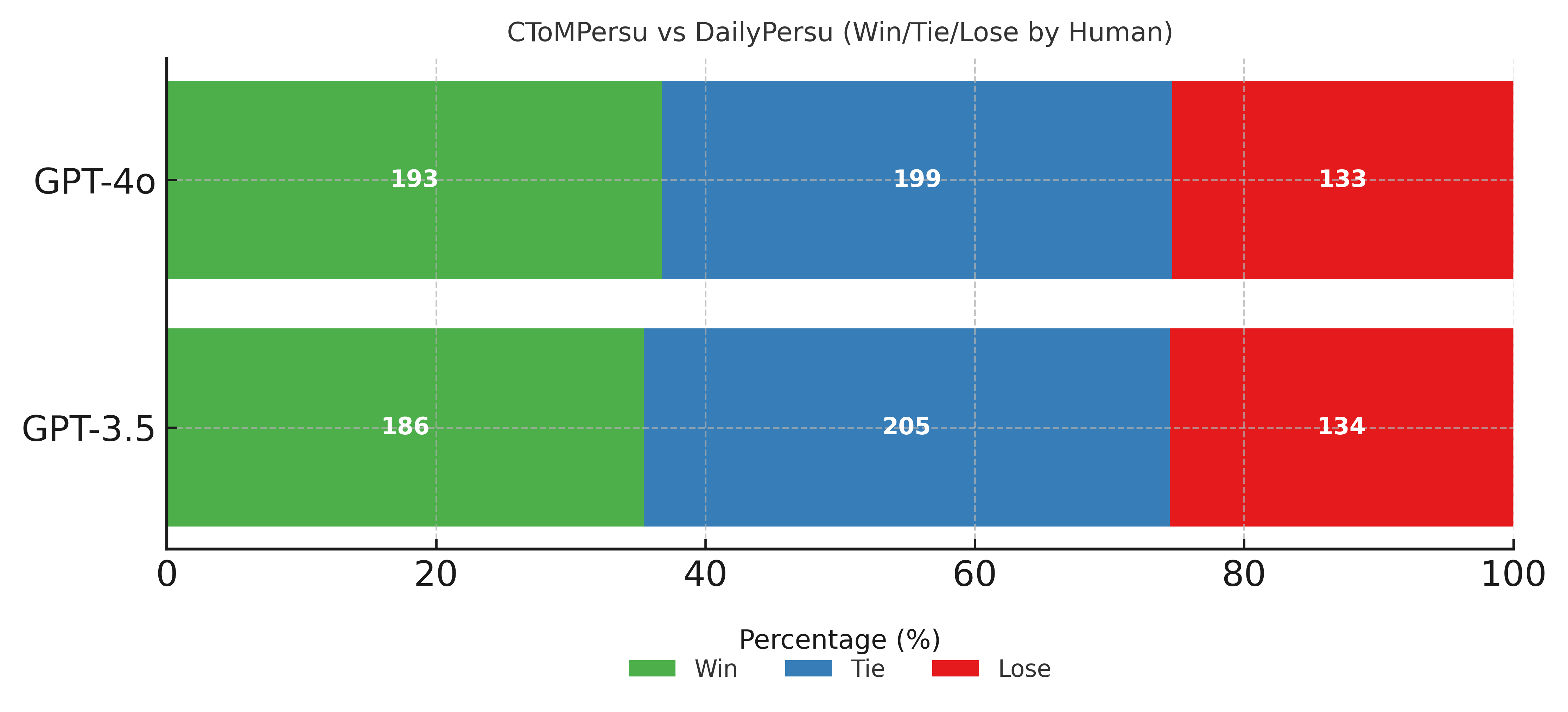}
        \vspace{0.3em}
        \small (b) Human Evaluation Results: CToMPersu vs DailyPersuasion
        \label{fig:output_human}
    \end{minipage}

    \caption{Comparison of LLM and human win/tie/lose evaluations between CToMPersu and DailyPersuasion.}
    \label{fig:output_combined}
    \vspace{-0.2cm}
\end{figure*}

\begin{figure}[t]
    \centering
    \includegraphics[width=0.485\textwidth]{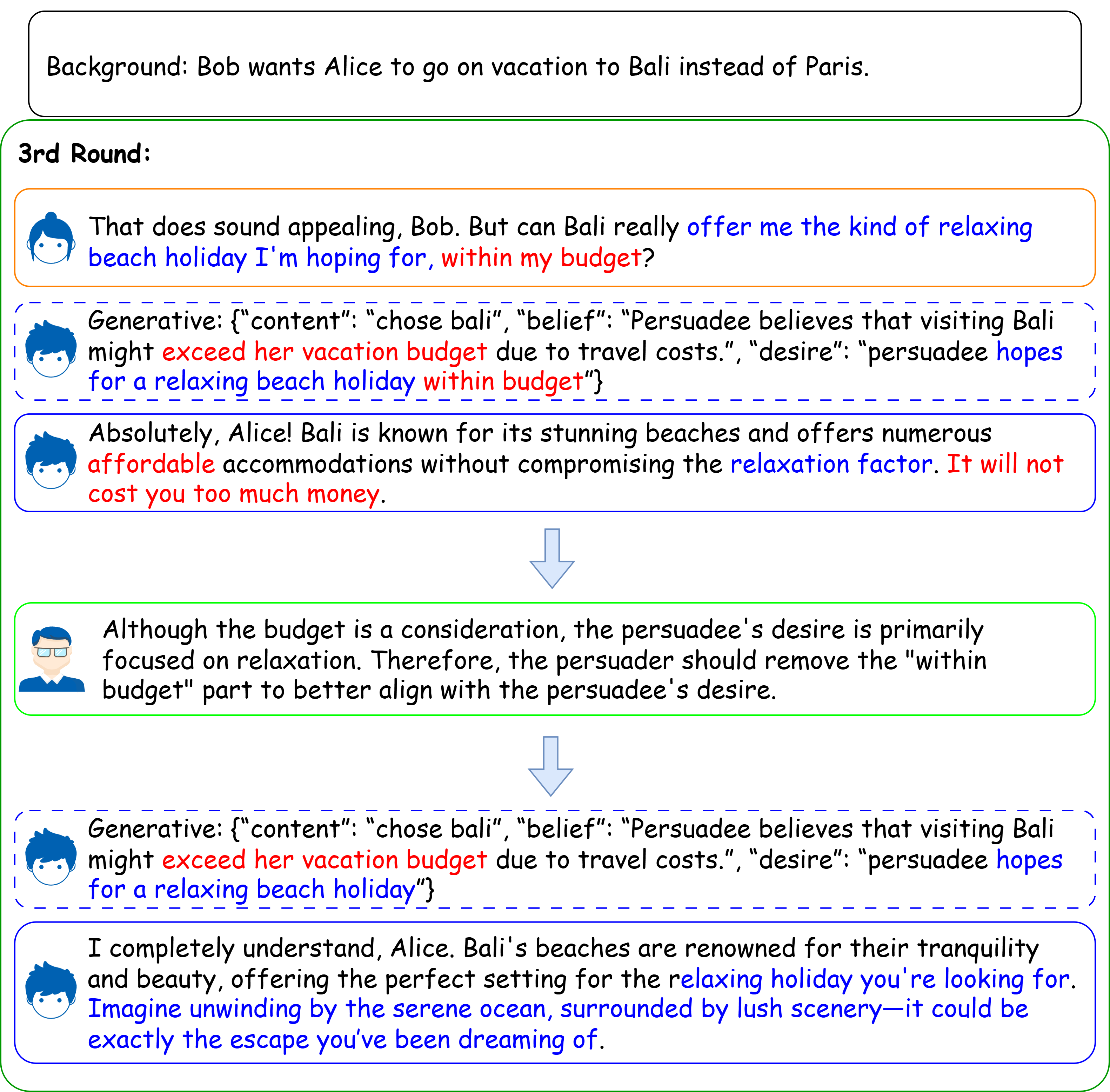}
    \caption{An example shows the observer agent’s effectiveness: the persuader should address the \textcolor{blue}{desire}, but both prediction and generation wrongly stress the \textcolor{red}{belief}, which the observer ultimately corrects.}
    \label{fig:observer}
    \vspace{-0.2cm}
\end{figure}

\noindent model performance in persuasive dialogue generation. In all settings, the persuadee was consistently played by GPT-4o, with access to scenario background, mental state annotations, and behavioral guidelines. In contrast, the persuader agent had no access to this information, preserving a double-blind condition throughout the interaction.

We examine the impact of different external knowledge sources on persuasive dialogue generation using an in-context learning (ICL) setup. For each test instance, the persuader agent retrieves same-domain dialogue examples from the training data as ICL demonstrations. To ensure fair comparison across datasets, only the dialogue content of the retrieved examples is provided, while scenario descriptions are excluded, since CToMPersu contains additional structured annotations (e.g., belief and desire states) that are not available in DailyPersuasion.

To assess persuasive effectiveness, we conduct both LLM-based and human evaluations. Human evaluation is performed by four graduate students in computer science using blind pairwise A/B comparison. For each instance, annotators are presented with two model-generated dialogues—one using CToMPersu demonstrations and the other using DailyPersuasion—without knowing the underlying data source, and are asked to judge which dialogue is more persuasive or whether they are equally persuasive. Final labels are determined by majority vote.

As shown in Figure~\ref{fig:output_combined}, results from both LLM-based and human evaluations consistently favor CToMPersu over DailyPersuasion. This suggests that CToMPersu provides more effective guidance when used as an external knowledge base, leading to more persuasive dialogue generation by the model.

\subsection{Observer Agent Case Study}

At times, the persuader agent may misjudge or make errors in predicting the persuadee agent's mental state. For instance, as illustrated in Figure~\ref{fig:observer}, the persuader was expected to address the persuadee’s desire regarding Generative Behavior, since the belief had already been resolved in the previous round. However, when the persuadee agent expressed their desire, it included the phrase "within my budget," which corresponded to a belief that had already been addressed. The true desire, however, was simply "hope for relaxation." As a result, the persuader agent mistakenly incorporated the budget constraint into their assessment of the desire, leading to a response that overly focused on the budget. This diminished the effectiveness of the persuasion, as the response should have primarily addressed the "relaxation" aspect. Ultimately, with guidance from the Observer Agent, the persuader corrected their prediction of the desire and generated a more targeted response, avoiding unnecessary discussion about the budget.

\section{Conclusion}
In this work, we address key challenges in developing AI-driven persuasion systems that more closely align with real human dialogue dynamics. We introduce a novel evaluation method based on causal theory of mind, enabling the LLM to infer and address the persuadee’s beliefs and desires. Through the development of ToMMA, a multi-agent framework, we ensure double-blind conditions and guide persuasive dialogues with causal reasoning, leading to more human-like interactions. Additionally, we present CToMPersu, a large-scale, multi-domain dataset that effectively addresses logical inconsistencies and demonstrates strong alignment with human dialogues, marking a significant advancement in realistic persuasive dialogue generation.

\section*{Compliance with Ethical Standards}
This study involved human evaluation of model outputs. 
No personally identifiable information was collected and the task involved minimal risk to participants. 
In accordance with institutional guidelines, ethical approval was not required for this type of study. 
The authors declare no conflicts of interest.

\bibliographystyle{IEEEbib}
\bibliography{strings, custom, refs}

@article{rogiers2024persuasion,
  title={Persuasion with Large Language Models: a Survey},
  author={Rogiers, Alexander and Noels, Sander and Buyl, Maarten and De Bie, Tijl},
  journal={arXiv preprint arXiv:2411.06837},
  year={2024}
}

@inproceedings{10.5555/3327546.3327641,
author = {Li, Raymond and Kahou, Samira and Schulz, Hannes and Michalski, Vincent and Charlin, Laurent and Pal, Chris},
title = {Towards deep conversational recommendations},
year = {2018},
publisher = {Curran Associates Inc.},
booktitle = ProcNeu,
pages = {9748–9758},
numpages = {11},
location = {Montr\'{e}al, Canada}
}

@article{altay2023information,
  title={Information delivered by a chatbot has a positive impact on COVID-19 vaccines attitudes and intentions.},
  author={Altay, Sacha and Hacquin, Anne-Sophie and Chevallier, Coralie and Mercier, Hugo},
  journal={Journal of Experimental Psychology: Applied},
  volume={29},
  number={1},
  pages={52},
  year={2023},
  publisher={American Psychological Association}
}

@article{chen2023would,
  title={Would an AI chatbot persuade you: an empirical answer from the elaboration likelihood model},
  author={Chen, Qian and Yin, Changqin and Gong, Yeming},
  journal={Information Technology \& People},
  year={2023},
  publisher={Emerald Publishing Limited}
}

@article{premack1978does,
  title={Does the chimpanzee have a theory of mind?},
  author={Premack, David and Woodruff, Guy},
  journal={Behavioral and brain sciences},
  volume={1},
  number={4},
  pages={515--526},
  year={1978},
  publisher={Cambridge University Press}
}

@inproceedings{georgeff1999belief,
  title={The belief-desire-intention model of agency},
  author={Georgeff, Michael and Pell, Barney and Pollack, Martha and Tambe, Milind and Wooldridge, Michael},
  booktitle={Proc. ATAL},
  pages={1--10},
  year={1999},
  organization={Springer}
}

@inproceedings{wu2024change,
  title={How to Change a Mind: Adults and Children Use the Causal Structure of Theory of Mind to Intervene on Others’ Behaviors},
  author={Wu, Shengyi and Schulz, Laura and Saxe, Rebecca},
  booktitle={Proceedings of the Annual Meeting of the Cognitive Science Society},
  volume={46},
  year={2024}
}

@article{baron1985does,
  title={Does the autistic child have a “theory of mind”?},
  author={Baron-Cohen, Simon and Leslie, Alan M and Frith, Uta},
  journal={Cognition},
  volume={21},
  number={1},
  pages={37--46},
  year={1985},
  publisher={Elsevier}
}

@article{yim2024evaluating,
  title={Evaluating and enhancing llms agent based on theory of mind in guandan: A multi-player cooperative game under imperfect information},
  author={Yim, Yauwai and Chan, Chunkit and Shi, Tianyu and Deng, Zheye and Fan, Wei and Zheng, Tianshi and Song, Yangqiu},
  journal={arXiv preprint arXiv:2408.02559},
  year={2024}
}

@article{cross2024hypothetical,
  title={Hypothetical minds: Scaffolding theory of mind for multi-agent tasks with large language models},
  author={Cross, Logan and Xiang, Violet and Bhatia, Agam and Yamins, Daniel LK and Haber, Nick},
  journal={arXiv preprint arXiv:2407.07086},
  year={2024}
}

@article{lou2025personalitymodelingpersuasionmisinformation,
      title={Personality Modeling for Persuasion of Misinformation using AI Agent}, 
      author={Qianmin Lou and Wentao Xu},
      journal={arXiv preprint arXiv:2501.08985},
      year={2025},
      eprint={2501.08985},
      archivePrefix={arXiv},
      primaryClass={cs.CL},
      url={https://arxiv.org/abs/2501.08985}, 
}

@inproceedings{jin-etal-2023-joint,
    title = "Joint Semantic and Strategy Matching for Persuasive Dialogue",
    author = "Jin, Chuhao  and
      Zhu, Yutao  and
      Kong, Lingzhen  and
      Li, Shijie  and
      Zhang, Xiao  and
      Song, Ruihua  and
      Chen, Xu  and
      Chen, Huan  and
      Sun, Yuchong  and
      Chen, Yu  and
      Xu, Jun",
    booktitle = FindEMNLP,
    month = dec,
    year = "2023",
    publisher = "Association for Computational Linguistics",
    url = "https://aclanthology.org/2023.findings-emnlp.276/",
    doi = "10.18653/v1/2023.findings-emnlp.276",
    pages = "4187--4197"
}

@inproceedings{wang-etal-2019-persuasion,
    title = "Persuasion for Good: Towards a Personalized Persuasive Dialogue System for Social Good",
    author = "Wang, Xuewei  and
      Shi, Weiyan  and
      Kim, Richard  and
      Oh, Yoojung  and
      Yang, Sijia  and
      Zhang, Jingwen  and
      Yu, Zhou",
    booktitle = ProcACL,
    month = jul,
    year = "2019",
    publisher = "Association for Computational Linguistics",
    url = "https://aclanthology.org/P19-1566/",
    doi = "10.18653/v1/P19-1566",
    pages = "5635--5649"
}

@inproceedings{zeng-etal-2020-meddialog,
    title = "{M}ed{D}ialog: Large-scale Medical Dialogue Datasets",
    author = "Zeng, Guangtao  and
      Yang, Wenmian  and
      Ju, Zeqian  and
      Yang, Yue  and
      Wang, Sicheng  and
      Zhang, Ruisi  and
      Zhou, Meng  and
      Zeng, Jiaqi  and
      Dong, Xiangyu  and
      Zhang, Ruoyu  and
      Fang, Hongchao  and
      Zhu, Penghui  and
      Chen, Shu  and
      Xie, Pengtao",
    booktitle = ProcEMNLP,
    month = nov,
    year = "2020",
    publisher = "Association for Computational Linguistics",
    url = "https://aclanthology.org/2020.emnlp-main.743/",
    doi = "10.18653/v1/2020.emnlp-main.743",
    pages = "9241--9250"
}

@inproceedings{jin-etal-2024-persuading,
    title = "Persuading across Diverse Domains: a Dataset and Persuasion Large Language Model",
    author = "Jin, Chuhao  and
      Ren, Kening  and
      Kong, Lingzhen  and
      Wang, Xiting  and
      Song, Ruihua  and
      Chen, Huan",
    booktitle = ProcACL,
    month = aug,
    year = "2024",
    publisher = "Association for Computational Linguistics",
    url = "https://aclanthology.org/2024.acl-long.92/",
    doi = "10.18653/v1/2024.acl-long.92",
    pages = "1678--1706"
}

@inproceedings{potter-etal-2024-hidden,
    title = "Hidden Persuaders: {LLM}s' Political Leaning and Their Influence on Voters",
    author = "Potter, Yujin  and
      Lai, Shiyang  and
      Kim, Junsol  and
      Evans, James  and
      Song, Dawn",
    booktitle = ProcEMNLP,
    month = nov,
    year = "2024",
    publisher = "Association for Computational Linguistics",
    url = "https://aclanthology.org/2024.emnlp-main.244/",
    doi = "10.18653/v1/2024.emnlp-main.244",
    pages = "4244--4275"
}

@inproceedings{saenger-etal-2024-autopersuade,
    title = "{A}uto{P}ersuade: A Framework for Evaluating and Explaining Persuasive Arguments",
    author = "Saenger, Till Raphael  and
      Hinck, Musashi  and
      Grimmer, Justin  and
      Stewart, Brandon M.",
    booktitle = ProcEMNLP,
    month = nov,
    year = "2024",
    publisher = "Association for Computational Linguistics",
    url = "https://aclanthology.org/2024.emnlp-main.913/",
    doi = "10.18653/v1/2024.emnlp-main.913",
    pages = "16325--16342"
}

@inproceedings{le-etal-2019-revisiting,
    title = "Revisiting the Evaluation of Theory of Mind through Question Answering",
    author = "Le, Matthew  and
      Boureau, Y-Lan  and
      Nickel, Maximilian",
    booktitle = ProcEMNLPIJCNLP,
    month = nov,
    year = "2019",
    publisher = "Association for Computational Linguistics",
    url = "https://aclanthology.org/D19-1598/",
    doi = "10.18653/v1/D19-1598",
    pages = "5872--5877"
}

@inproceedings{kim-etal-2023-fantom,
    title = "{FANT}o{M}: A Benchmark for Stress-testing Machine Theory of Mind in Interactions",
    author = "Kim, Hyunwoo  and
      Sclar, Melanie  and
      Zhou, Xuhui  and
      Bras, Ronan  and
      Kim, Gunhee  and
      Choi, Yejin  and
      Sap, Maarten",
    booktitle = ProcEMNLP,
    month = dec,
    year = "2023",
    publisher = "Association for Computational Linguistics",
    url = "https://aclanthology.org/2023.emnlp-main.890/",
    doi = "10.18653/v1/2023.emnlp-main.890",
    pages = "14397--14413"
}

@inproceedings{chen-etal-2024-tombench,
    title = "{T}o{MB}ench: Benchmarking Theory of Mind in Large Language Models",
    author = "Chen, Zhuang  and
      Wu, Jincenzi  and
      Zhou, Jinfeng  and
      Wen, Bosi  and
      Bi, Guanqun  and
      Jiang, Gongyao  and
      Cao, Yaru  and
      Hu, Mengting  and
      Lai, Yunghwei  and
      Xiong, Zexuan  and
      Huang, Minlie",
    booktitle = ProcACL,
    month = aug,
    year = "2024",
    publisher = "Association for Computational Linguistics",
    url = "https://aclanthology.org/2024.acl-long.847/",
    doi = "10.18653/v1/2024.acl-long.847",
    pages = "15959--15983"
}

@article{ju2025adaptivepsychologicalpersuasionlarge,
      title={On the Adaptive Psychological Persuasion of Large Language Models}, 
      author={Tianjie Ju and Yujia Chen and Hao Fei and Mong-Li Lee and Wynne Hsu and Pengzhou Cheng and Zongru Wu and Zhuosheng Zhang and Gongshen Liu},
      journal={arXiv preprint arXiv:2506.06800},
      year={2025},
      eprint={2506.06800},
      archivePrefix={arXiv},
      primaryClass={cs.CL},
      url={https://arxiv.org/abs/2506.06800}, 
}

@article{moore2025largelanguagemodelsplanning,
      title={Do Large Language Models Have a Planning Theory of Mind? Evidence from MindGames: a Multi-Step Persuasion Task}, 
      author={Jared Moore and Ned Cooper and Rasmus Overmark and Beba Cibralic and Nick Haber and Cameron R. Jones},
      journal={arXiv preprint arXiv:2507.16196},
      year={2025},
      eprint={2507.16196},
      archivePrefix={arXiv},
      primaryClass={cs.CL},
      url={https://arxiv.org/abs/2507.16196}, 
}

@inproceedings{ivanov2024checkworthy,
  author    = {Ivanov, Petar and Koychev, Ivan and Hardalov, Momchil and Nakov, Preslav},
  title     = {Detecting Check-Worthy Claims in Political Debates, Speeches, and Interviews Using Audio Data},
  booktitle = {Proc. IEEE Int. Conf. Acoustics, Speech and Signal Processing (ICASSP)},
  year      = {2024},
  pages     = {12011--12015},
  doi       = {10.1109/ICASSP48485.2024.10447064}
}

@inproceedings{shi2024charismatic,
  author    = {Shi, Rongjie and Niebuhr, Oliver and Gu, Wentao and Taghva, Nasieh},
  title     = {The Effects of Loudness and Smiling on Timbre Features: Implications for Charismatic Voices in Mandarin, German and Danish},
  booktitle = {Proc. IEEE Int. Conf. Acoustics, Speech and Signal Processing (ICASSP)},
  year      = {2024},
  pages     = {11926--11930},
  doi       = {10.1109/ICASSP48485.2024.10446124}
}

@string{ProcEMNLP = "Proc. EMNLP"}

@string{FindEMNLP    = "Findings of EMNLP"}

@string{ProcEMNLPIJCNLP = "Proc. EMNLP-IJCNLP"}

@string{ProcACL = "Proc. ACL"}

@string{ProcNeu = "Proc. NeurIPS"}

\end{document}